\documentclass{article}

\usepackage[preprint]{neurips_2022}

\usepackage{mmstyles}

\usepackage{amsmath,amsfonts,bm}

\def\eqref#1{equation~(\ref{#1})}

\def\1{\bm{1}}

\def\vc{{\bm{c}}}

\def\vg{{\bm{g}}}

\def\vx{{\bm{x}}}
\def\vy{{\bm{y}}}
\def\vz{{\bm{z}}}

\def\mI{{\bm{I}}}

\DeclareMathAlphabet{\mathsfit}{\encodingdefault}{\sfdefault}{m}{sl}
\SetMathAlphabet{\mathsfit}{bold}{\encodingdefault}{\sfdefault}{bx}{n}

\def\gN{{\mathcal{N}}}

\newcommand{\pdata}{p_{\rm{data}}}

\usepackage[utf8]{inputenc} 
\usepackage[T1]{fontenc}    
\usepackage{hyperref}       
\usepackage{url}            
\usepackage{booktabs}       
\usepackage{amsfonts}       
\usepackage{nicefrac}       
\usepackage{microtype}      
\usepackage{xcolor}         
\usepackage{bm}
\usepackage{amsmath}
\usepackage{amssymb}
\usepackage{amsthm}
\usepackage{latexsym}   
\usepackage{bbding}
\usepackage{algorithm}
\usepackage{algorithmic}
\usepackage{hyperref}
\usepackage{booktabs}
\usepackage{multirow}
\usepackage{graphicx}
\usepackage{url}
\usepackage{authblk}                            
\usepackage{array}
\usepackage{booktabs}
\usepackage{xcolor}
\usepackage{mathrsfs}
\usepackage{diagbox}
\usepackage{subfigure}

\title{Guided Diffusion Model for Adversarial Purification}

\newcommand{\AuthorSpace}{\hspace{1.2em}}
\author{%
 \textbf{\thanks{Equal Contribution} \hspace{0.08em} Jinyi Wang$^{1,3}$\AuthorSpace{} $^*$ Zhaoyang Lyu$^{2,3}$\AuthorSpace{} Dahua Lin$^{2,3}$\AuthorSpace{} Bo Dai$^{3}$\AuthorSpace{} Hongfei Fu$^{1}$}\\
 $^1$Shanghai JiaoTong University \\
 $^2$The Chinese University of Hong Kong \\
 $^3$Shanghai AI Laboratory \\
 \texttt{jinyi.wang@sjtu.edu.cn, lyuzhaoyang@link.cuhk.edu.hk}, \\
\texttt{dhlin@ie.cuhk.edu.hk, daibo@pjlab.org.cn, fuhf@cs.sjtu.edu.cn}
}

\begin{document}

\maketitle

\begin{abstract}
With wider application of deep neural networks (DNNs) in various algorithms and frameworks, security threats have become one of the concerns. Adversarial attacks disturb DNN-based image classifiers, in which attackers can intentionally add imperceptible adversarial perturbations on input images to fool the classifiers.
In this paper,
we propose a novel purification approach,
referred to as \textbf{guided diffusion model for purification (GDMP)}, to help protect classifiers from adversarial attacks.
The core of our approach is to embed purification into the diffusion-denoising process of a Denoised Diffusion Probabilistic Model (DDPM), so that its diffusion process could submerge the adversarial perturbations with gradually added Gaussian noises,
and both of these noises can be simultaneously removed following a guided denoising process.
On our comprehensive experiments across various datasets, the proposed GDMP is shown to reduce the perturbations raised by adversarial attacks to a shallow range, 
thereby significantly improving the correctness of classification.
GDMP improves the robust accuracy by 5\%, obtaining 90.1\% under PGD attack on the CIFAR10 dataset.
Moreover,
GDMP achieves 70.94\% robustness on the challenging ImageNet dataset. 

\end{abstract}

\section{Introduction}
\label{sec:introduction}
Thanks to the advance of deep neural networks (DNNs),
the performance of DNN-based image classifiers keep improving in recent years (\citet{DBLP:conf/nips/KrizhevskySH12};\citet{DBLP:conf/cvpr/SzegedyLJSRAEVR15};\citet{DBLP:journals/corr/SimonyanZ14a};\citet{DBLP:conf/cvpr/HeZRS16}). 
Despite their promising results,
DNN-based image classifiers are known to be sensitive to adversarial attacks.
Specifically, by altering images with slight perturbations that are undetectable to humans,
one can mislead DNN-based image classifiers to give unexpected predictions,
causing safety issues.
Therefore, 
finding a robust strategy to lower the disturbance from adversarial attacks is of great necessity to DNN-based image classifiers, especially when they are applied to real-world applications. 

A variety of strategies have been proposed to make DNN-based image classifiers resistant to adversarial attacks.
\textit{Adversarial training}(\citet{madry2018towards};\citet{zhang2019theoretically}) and \textit{adversarial purification}(\citet{DBLP:journals/nn/SrinivasanRMMSN21};\citet{DBLP:conf/icml/YangZXK19};\citet{yoon2021adversarial};\citet{DBLP:conf/iclr/ShiHM21}) are two common approaches.
Adversarial training aims at involving adversarial instances or optimizing at worst-case examples in the training process of neural networks.
Although it improves the robustness of target classifiers over seen types of adversarial attacks,
it often falls short of handling unseen types of adversarial attacks or image corruptions.
Adversarial purification, on the other hand, tries to purify attacked images before sending them to target classifiers,
which is often achieved by learning generative models (\citet{DBLP:conf/iclr/SamangoueiKC18};\citet{DBLP:conf/iclr/SongKNEK18};\citet{DBLP:conf/aaai/GhoshLB19};\citet{DBLP:conf/iclr/SchottRBB19}) as a purification models to convert attacked images into clean ones.
The purification model is usually trained independently of target classifiers, and is not restricted to specific types of adversarial attacks.
In this way, adversarial purification leaves target classifiers unaffected,
while demonstrating strong generalization ability to purify not only various types of adversarial attacks,
but also image corruptions such as image blurs.
However, 
it remains a challenging task to obtain a good purification model,
especially on large-scale datasets such as ImageNet (\citet{imagenet_cvpr09}),
since there often exists a trade-off between faithfully retaining the semantic content of clean images and completely removing the perturbation from attacked images.


\begin{figure}[t]
	\centering
    \includegraphics[width=1\textwidth]{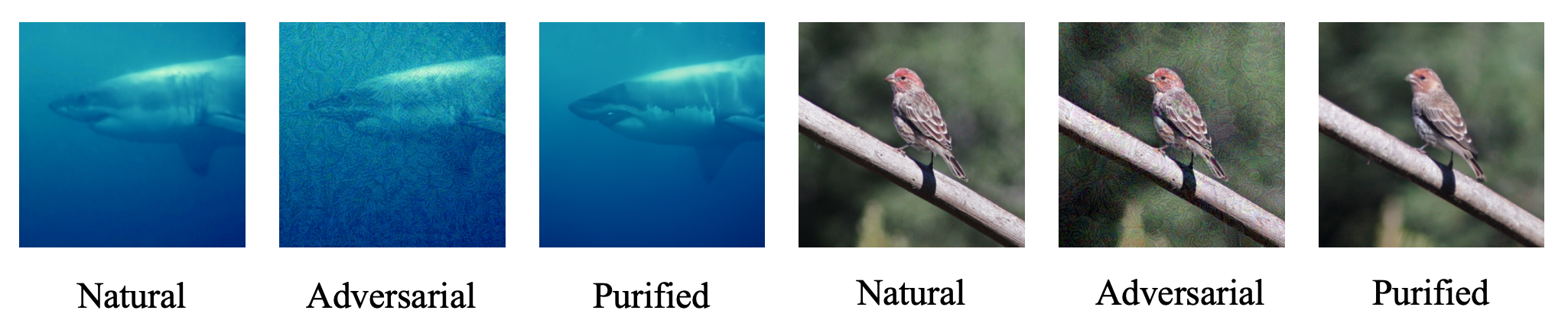}
    \vspace{-7mm}
	\caption{Adversarial samples with PGD attack $\ell_{\infty} \varepsilon$-ball with $\varepsilon$= 16/255 and the images after our purification method.}
	\vspace{-3mm}
	\label{fig:Effect} 
\end{figure}

Inspired by the success of Denoising Diffusion Probabilistic Model (DDPM) in image generation \citep{Sohl-Dickstein2015Deep,ho2020denoising},
in this work we propose a novel adversarial purification model based on DDPM.
Different from commonly used generative models such as GANs \citet{goodfellow2014generative} and VAEs~\citet{kingma2013auto},
DDPMs define a diffusion process that gradually adds noise to a clean image until it turns into a Gaussian noise.
To sample images, DDPMs use a neural network to simulate the reverse process of the diffusion process. 
The network iteratively denoises a Gaussian noise to a clean image.
We therefore can naturally embed adversarial purification into the diffusion-denoising process of DDPM.
Specifically,
we can use few diffusion steps to inject the attacked image with Gaussian noises to submerge its adversarial perturbations,
followed by applying the denoising process to eliminate both Gaussian noises and adversarial perturbations simultaneously.
To tackle the trade-off of adversarial purification 
and retaining image semantics, 
we propose to further equip the proposed approach with additional guidance during the denoising process to retain the consistency between the purified image and the original one.
The guidance is implemented by minimizing the distance (mean squared error) or maximizing the similarity (SSIM)(\citet{DBLP:journals/tip/WangBSS04} between the adversarial image and the denoised image.
This is because the adversarial image is close to the original clean image in terms of pixel values, and thus the denoised image will be close to the original clean image by encouraging it to be close to the adversarial image. 
We need to choose an appropriate guidance scale to enure that the denoised image is close to the adversarial image but with adversarial perturbations removed at the same time.

Compared to purification methods based on other generative models, our DDPM-based purification approach enjoys several important advantages.
First, we can only partially diffuse the adversarial image, and thus preserve most local details in the image. 
This is because the latent space of a DDPM has the same dimension of the original input image, and we can choose an appropriate diffusion time step to let the Gaussian noise submerge the adversarial perturbation, but at the same time preserve major contents in the input image.  
Moreover,
as shown in our experiments,
the novel guidance introduced to our approach can boost the purification performance when using a large number of diffusion steps,
leading to stronger purification effect.
Specifically,
we are the first to apply adversarial purification to a large-scale dataset such as ImageNet,
where we achieve state-of-the-art robustness accuracy (70.94\%) with the PGD attack ,
outperforming previous methods with a large margin.

\section{Related Work}

\paragraph{Adversarial Training.}
Consider a classifier neural network $g_{\phi}: \mathbb{R}^{D} \rightarrow \mathbb{R}^{K}$, where $D$ represents the dimension of input data and $K$ represents the number of class labels. 
Worst-case risk minimization can be used to train a neural network resistant to adversarial attacks, 
\begin{equation}
\label{eqn:advtraining}
\min _{\phi} \mathbb{E}_{p_{\text {data }}(\vx, \vy)}\left[\max _{\vx^{\prime} \in \mathcal{B}(\vx)} \mathcal{L}\left(g_{\phi}\left(\vx^{\prime}\right), \vy\right)\right],
\end{equation}

where $\mathcal{L}$ is a loss function, and $\mathcal{B}(\vx)$ is the set of allowed perturbed images around $\vx$. 
An approximation must be used because perfect evaluation and optimization of this goal are impossible. 
Adversarial training
\citep{madry2018towards, zhang2019theoretically, carmon2019unlabeled}
is a popular method for optimizing a surrogate loss computed from limited classes of adversarial samples during training.
\citet{madry2018towards}, for example, uses the PGD attack to approximate the inner maximizing loop in Equation~\ref{eqn:advtraining}.

\begin{equation}
\min _{\phi} \mathbb{E}_{p_{\text {data }}(\vx, \vy)}\left[\mathbb{E}_{\vx^{\prime} \in \mathcal{A}(\vx)} \mathcal{L}\left(g_{\phi}\left(\vx^{\prime}\right), \vy\right)\right],
\end{equation}
where $\mathcal{A}(\vx_i)$ denotes the set of images obtained on the fly in the training process of the network by hitting $\vx$ with a certain type of adversarial attack. 

\paragraph{Prepossessing.}
Preprocessing, in which input images are preprocessed using auxiliary transformations before classification, is another strategy for adversarial defense. Let $f_{\theta}$ be a preprocessor for image transformation.
The goal of the training is then:
\begin{equation}
\min _{\phi, \theta} \mathbb{E}_{p_{\text {data }}(\vx, \vy)}\left[\max _{\vx^{\prime} \in \mathcal{B}(\vx)} \mathcal{L}\left(g_{\phi}\left(f_{\theta}\left(\vx^{\prime}\right)\right), \vy\right)\right].
\end{equation}
An average over recognized types of adversarial attacks or an average over stochastically altered inputs can be used to approximate the maximum over the threat model $\mathcal{B}(\vx)$. Adding stochasticity to the input images, as well as discontinuous or non-differentiable transforms (\citet{DBLP:conf/iclr/GuoRCM18};\citet{DBLP:conf/iclr/DhillonALBKKA18};\citet{DBLP:conf/iclr/BuckmanRRG18};\citet{DBLP:conf/iclr/XiaoZZ20}), makes gradient estimation concerning the loss function $\nabla_\vx \mathcal{L}\left(g_{\phi}\left(f_{\theta}\left(\vx^{\prime}\right)\right), \vy\right)$ harder to attack.

\paragraph{Adversarial purification.}
Finally, the adversarial purification is one of the Prepossessing methods.
A generative model that can restore clean images from attacked images is trained and used as a preprocessor in adversarial purification (\citet{DBLP:conf/iclr/SamangoueiKC18};\citet{DBLP:conf/iclr/SongKNEK18}; \citet{DBLP:journals/nn/SrinivasanRMMSN21};\citet{DBLP:conf/aaai/GhoshLB19}; \citet{DBLP:conf/iclr/HillMZ21}; \citet{DBLP:conf/iclr/ShiHM21};\citet{yoon2021adversarial}), where the preprocessor $f$ corresponds to the purification. 

\citet{DBLP:conf/iclr/SamangoueiKC18} proposed defense-GAN, a generator that can restore clean images from attacked images. \citet{DBLP:conf/iclr/SongKNEK18} revealed how to detect and purify adversarial samples using an autoregressive generative model. \citet{DBLP:journals/nn/SrinivasanRMMSN21} presented a purification approach for denoising autoencoders using the Metropolis Adjusted Langevin algorithm (MALA).\citet{DBLP:conf/iclr/GrathwohlWJD0S20},\citet{DBLP:conf/nips/DuM19} and \citet{DBLP:conf/iclr/HillMZ21} proposed and demonstrated that MCMC wich EBMs can purify adversarial examples. Similarly, \citet{yoon2021adversarial} used the denoising score-based model for purification.

\section{Background of DDPM}
\subsection{DDPM }
\newcommand{\platent}{p_{\rm{latent}}}
\label{sec:diffusion_model}
Assume $\pdata$ is the distribution of all images $\vx$ in the dataset, and $\platent=\gN(\bm{0}_{3N},\mI_{3N\times3N})$ is the latent distribution, and that $\gN$ is the Gaussian distribution. 
DDPMs consist of two Markov chains, the diffusion process and the reverse process. 
The diffusion process gradually adds noise to a clean image until it becomes a Gaussian noise, 
and the reverse process iteratively denoises a sampled Gaussian noise to a clean image. 
The length of both processes is $T$.

\paragraph{The Diffusion Process.}
The diffusion process is a Markov process that gradually adds noise to a clean image until it approaches $\platent$. 
Formally, let $\vx^0\sim\pdata$. 
The diffusion step $t$ is denoted by a superscript. 
The diffusion process from clean data $\vx^0$ to $\vx^T$ is defined as 
\begin{equation}
\label{eqn:diffusion_process}
    q(\vx^1,\cdots,\vx^T|\vx^0) = \prod_{t=1}^T q(\vx^t|\vx^{t-1}),
    \text{ where }
    q(\vx^t|\vx^{t-1})=\gN(\vx^t;\sqrt{1-\beta_t}\vx^{t-1},\beta_t \mI),
\end{equation}
where $\beta_t$'s are predefined small positive constants. We give the detailed definition of $\beta_t$ in the corresponding experiment part.
 According to \citet{ho2020denoising}, there is a closed form expression for $q(\vx^t|\vx^0)$. We first define constants $\alpha_t = 1 - \beta_t$, $\bar{\alpha}_t = \prod_{i=1}^t\alpha_i$. Then, we have
$q(\vx^t|\vx^0) = \gN\left(\vx^t; \sqrt{\bar{\alpha}_t}\vx^0, (1-\bar{\alpha}_t)\mI\right)$.

Therefore, when $T$ is large enough, $\bar{\alpha}_t$ goes to $0$, and $q(\vx^T|\vx^0)$ becomes close to the latent distribution $\platent(\vx^T)$.
Note that $\vx^t$ can be directly sampled through the following equation:
\begin{equation}
\label{eqn:xt|x0}
\vx^t = \sqrt{\bar{\alpha}_t}\vx^0 + \sqrt{1-\bar{\alpha}_t}\bm{\epsilon},
\text{ where $\vepsilon$ is a standard Gaussian noise.}
\end{equation}

\paragraph{The Reverse Process.}
The reverse process is a Markov process that predicts and eliminates the noise added in the diffusion process. Let $\vx^T\sim\platent$ be a latent variable. The reverse process from latent $\vx^T$ to clean data $\vx^0$ is defined as
\begin{align}
\label{eqn:reverse_process}
    p_{\bm{\theta}}(\vx^0,\cdots,\vx^{T-1}|\vx^T)=\prod_{t=1}^T p_{\bm{\theta}}(\vx^{t-1}|\vx^t),
    \text{ where }
    p_{\bm{\theta}}(\vx^{t-1}|\vx^t ) = \gN(\vx^{t-1};\bm{\mu}_{\bm{\theta}}(\vx^t, t), \sigma_t^2\mI).
\end{align}
The mean $\bm{\mu}_{\bm{\theta}}(\vx^t, t)$ is a neural network parameterized by $\bm{\theta}$, and the variance $\sigma_t^2$'s can be either time-step dependent constants~\citep{ho2020denoising} or learned by a neural network~\citep{nichol2021improved}. 
To generate a sample, we first sample $\vx^T \sim \gN(\bm{0}_{3N},\mI_{3N\times3N})$, then draw $\vx^{t-1}\sim p_{\bm{\theta}}(\vx^{t-1}|\vx^t)$ for $t=T,T-1,\cdots,1$, and finally outputs $\vx^0$.

\section{Methodology}

\label{sec:method}
\subsection{Diffusion Method for Adversarial Puriﬁcation}
We propose to use a pre-trained DDPM to purify adversarial samples.

The DDPM contains two Markov chains called the diffusion process and the reverse process, which adds Gaussian noise to a clean image and eliminates noises in the image, respectively. 

We can utilize these two processes to destroy and remove adversarial perturbations in an image.
The diffusion process can inject an adversarial image with Gaussian noises to submerge the adversarial perturbation. 
The reverse process can eliminate the Gaussian noises and the adversarial perturbation at the same time, and thus move the adversarial image to the domain of normal clean images. 

Assume that an adversarial image $\vx_{\text{adv}}$ contains adversarial perturbation as $\vx_{\text{adv}}=\vx + \delta$.
We diffuse the adversarial image $\vx_{\text{adv}}$ for $T_c$ steps following Equation~\ref{eqn:xt|x0}:

\begin{equation}
\label{eqn:adversarial perturbation}
    \vx^{T_c} = \sqrt{\bar{\alpha}_{T_c}}(\vx + \bm{\delta}) + \sqrt{1-\bar{\alpha}_{T_c}}\bm{\epsilon} = \sqrt{\bar{\alpha}_{T_c}}\vx + \sqrt{\bar{\alpha}_{T_c}}\bm{\delta} + \sqrt{1-\bar{\alpha}_{T_c}}\bm{\epsilon}.
\end{equation}

As ${T_c}$ increases, $\bar{\alpha}_{T_c}$ gradually decreases and $1-\bar{\alpha}_{T_c}$ gradually increases. 
We know that the norm of $\bm{\delta}$ is small compared with $\vx$ due to the perceptually indistinguishable constraint.
Therefore, we need to choose an appropriate ${T_c}$ such that the magnitude of the Gaussian noise $\sqrt{1-\bar{\alpha}_{T_c}}\bm{\epsilon}$ is large enough to submerge the adversarial perturbation $\sqrt{\bar{\alpha}_{T_c}} \delta$, while at the same time, the content in $\sqrt{\bar{\alpha}_{T_c}} \vx$ is largely preserved.
After diffusing the adversarial image, we can feed $\vx^{T_c}$ to the reverse process in Equation~\ref{eqn:reverse_process}.
The reverse process will eliminate the added Gaussian noise in $\vx^{T_c}$, and at the same time, very likely to remove the adversarial perturbation in $\vx^{T_c}$.
This is because for large enough ${T_c}$'s, the adversarial perturbation will be submerged and destroyed by the added Gaussian noise, $\vx^{T_c}$ will contain very little information of the adversarial perturbation $\delta$. 
Another reason is that DDPM as a generative model trained on normal clean images, it has the tendency to move noisy images $\vx^{T_c}$ to the domain of normal clean images in its reverse process.


\begin{figure*}[tb]
	\centering
	  \subfigure[The diffusion process and reverse process]{
       \centering
        \includegraphics[width=0.32\textwidth]{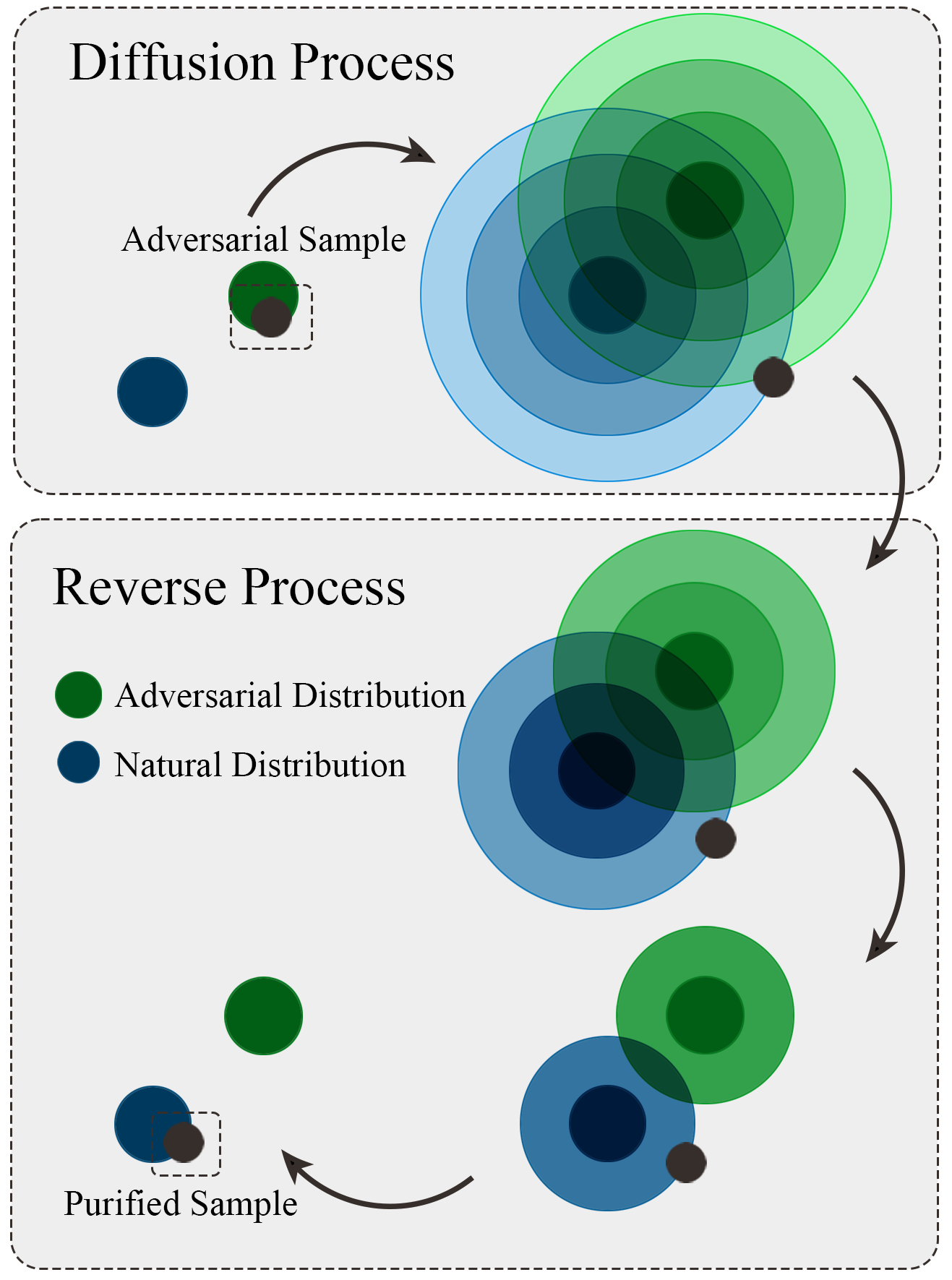}
    }
    \subfigure[How we add guidance during reverse process to avoid the randomness of reverse process.]{
        \centering
        \includegraphics[width=0.64\textwidth]{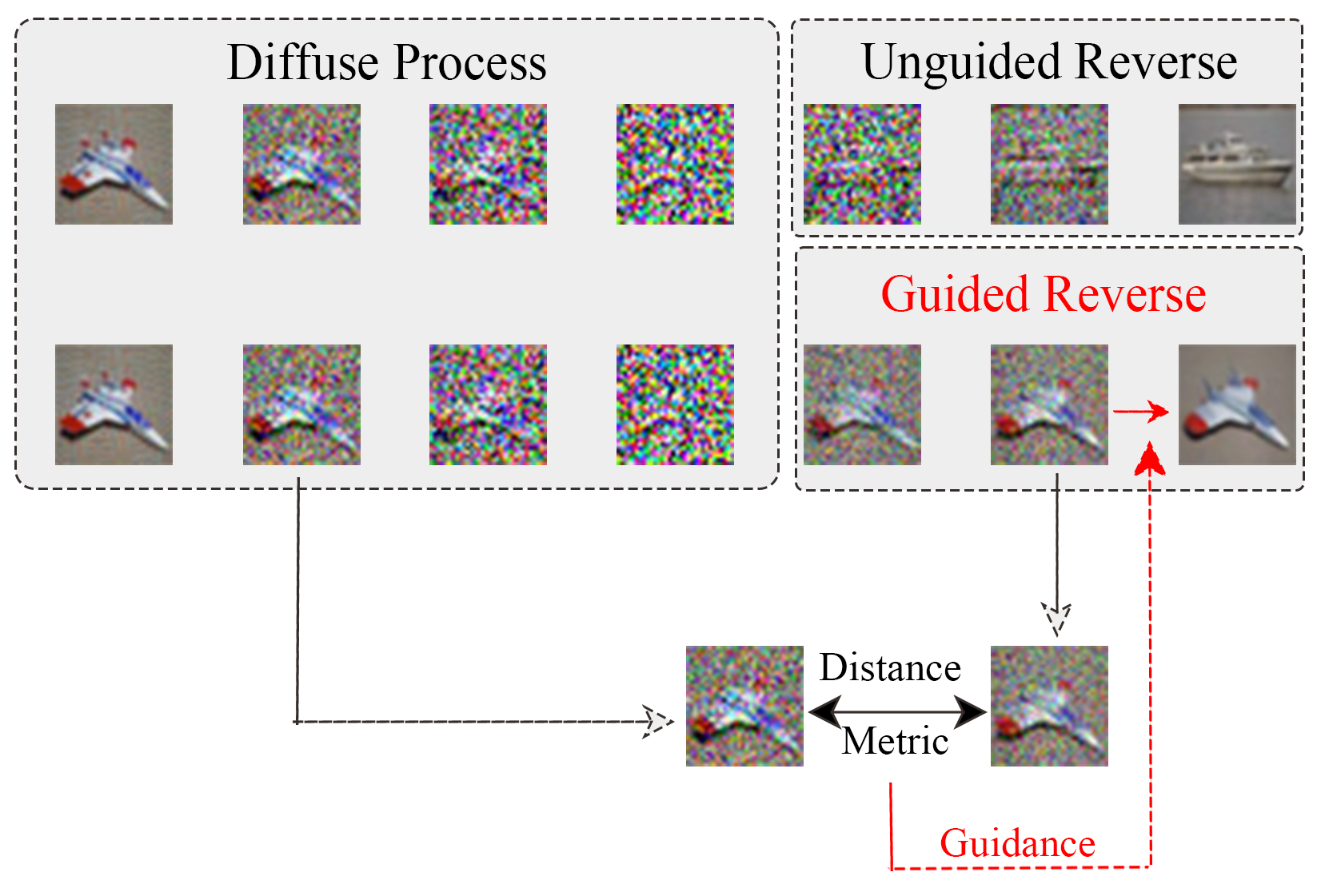}
    }
	\caption{This figure plots how the purification progress works. The black dots illustrate the purification  progress of our method. The \textcolor{green}{green} and \textcolor{blue}{blue} contour plots represent the distributions of adversarial images and  natural images. Given an adversarial image, we first diffuse it with Gaussian noise and remove the noise during reverse process steply. This progress gradually projects an adversarial image to a classifiable natural image. }
	\label{fig:diff_reason} 
\end{figure*}
In the above purification process, we can not choose a too large ${T_c}$, because we need to preserve the content in the clean image, but small ${T_c}$'s leads to weak purification effect. Some of the adversarial perturbation may remain in the purified image when ${T_c}$ is small.
To tackle this problem, we find that we can iteratively purify the adversarial images multiple times, each iteration with a not too large ${T_c}$. 
In this way, adversarial perturbation survived in previous purification iterations, could be removed in later iterations.
We find that this strategy is better than purifying the adversarial images once with a large ${T_c}$ in our experiments, which is shown in figure~\ref{fig:acc_step}(up).
The detailed purification algorithm with multiple iterations is listed in algorithm ~\ref{alg1:diffusion_reverse_process}.

\begin{algorithm}[htbp]
	\renewcommand{\algorithmicrequire}{\textbf{Input:}}
	\renewcommand{\algorithmicensure}{\textbf{Output:}}
	\caption{Adversarial purification with DDPM }
	\label{alg1:diffusion_reverse_process}
	\begin{algorithmic}[1]
		\REQUIRE an input $x$, diffusion length $T_c$ per each purification run, number of purification iterations M.
		\FOR {$i$ $\leftarrow 1$ to $M$}
            \STATE The diffusion process: $\vx^t = \sqrt{\bar{\alpha}_t}(x + v_{perb}) + \sqrt{1-\bar{\alpha}_t}\bm{\epsilon},$
            \FOR{$t \leftarrow T_c$ to $1$}
                \STATE The reverse process: $\vx^{t-1}\sim p_{\bm{\theta}}(\vx^{t-1}|\vx^t, \vc)$ 
            \ENDFOR
		\ENDFOR
		\RETURN $x_0$
	\end{algorithmic}
\end{algorithm}

\subsection{Guided DDPM}
\begin{figure*}[t] 
    \begin{minipage}[t]{0.60\textwidth}
        \centering
        \includegraphics[width=0.95\textwidth]{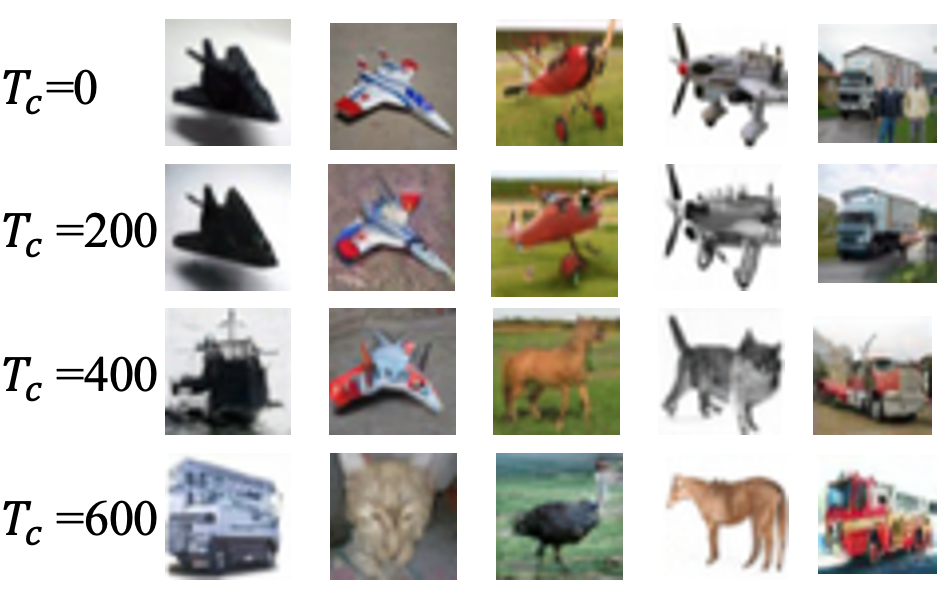}
        \caption{Generated images from $T_c$=0, 200, 400 and 600 steps diffusion and reverse process.}
        \label{fig:reverse_vary}
    \end{minipage}
    \begin{minipage}[t]{0.37\textwidth}
        \centering
        \includegraphics[width=0.98\textwidth]{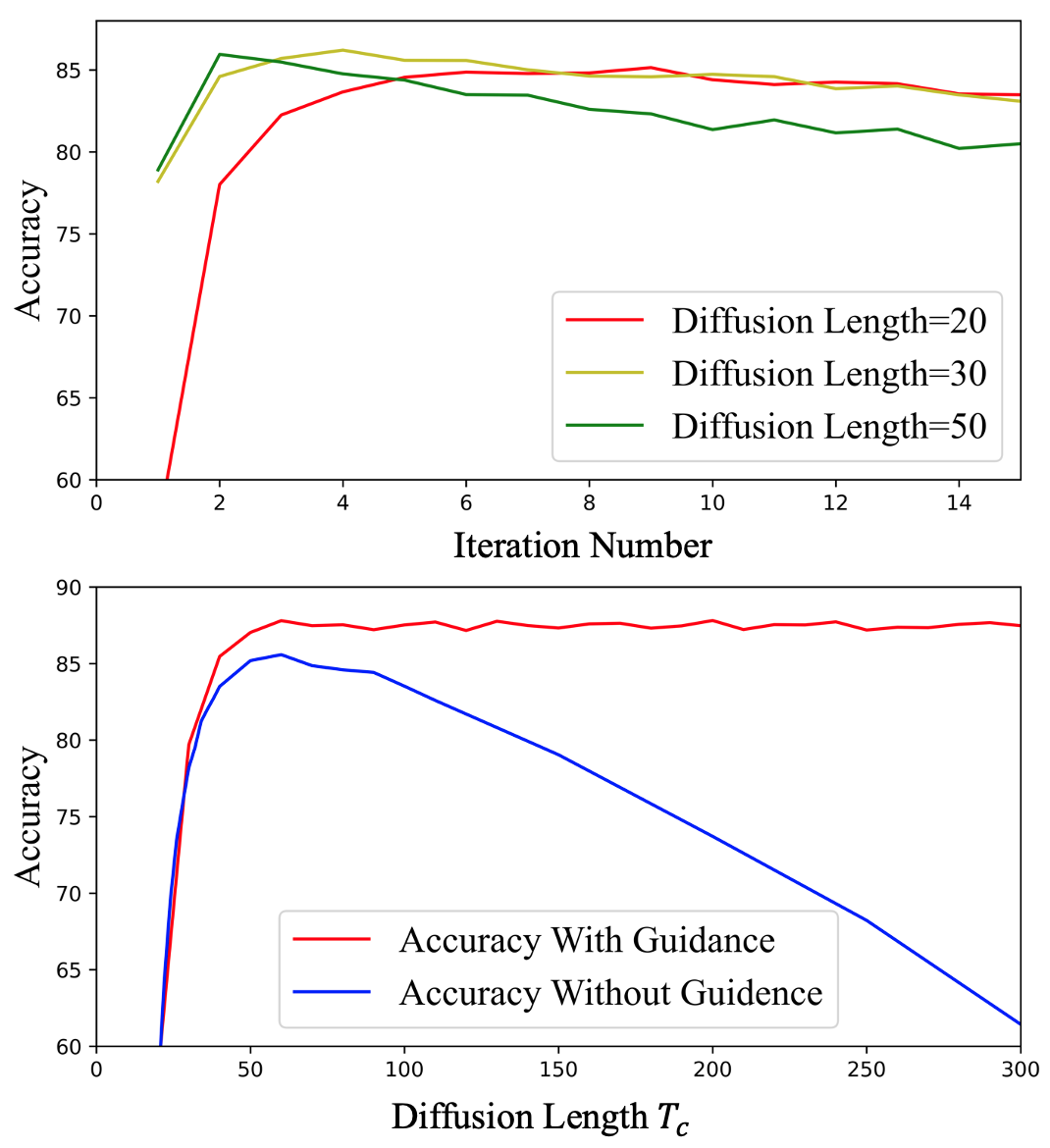}
        \caption{(up) Accuracy versus number of iteration on CIFAR10 dataset. 
        (down)Accuracy versus purification length with and without guidance on CIFAR10 dataset.}
        \label{fig:acc_step}
    \end{minipage}
\end{figure*}

As discussed in the above section, we face a trade-off between the purification effect and consistency with the original clean image.
If we use a too large diffusion length $T_c$, the purified images will deviate from the original clean image as shown in Figure~\ref{fig:reverse_vary}. 
Even if we use a small diffusion length with multiple purification iterations as described in the above section, the purified image will still deviate from the original clean image as the iteration number increases. As shown in Figure~\ref{fig:acc_step}, the classification accuracy would drop down when the iteration number become bigger. 
We need to design a method that enables us to use a large diffusion length $T_c$ to achieve strong purification effect, while at the same time ensures that the purified image is close the original clean image. 

Inspired by the work~\citep{DBLP:conf/nips/DhariwalN21}, in which the authors use a classifier to guide the reverse process of a DDPM, we propose to use the adversarial image as guidance in the reverse process.
We encourage the purified image to be close to the adversarial image in the reverse process of the DDPM.
This is because the adversarial image is close to the original clean image in terms of pixel values, and thus the purified image will be close to the original clean image by encouraging it to be close to the adversarial image. 
We need to choose an appropriate guidance scale, so that the purified image is close to the adversarial image, and thus the original clean image, 
but at the same time, the guidance scale should not be too large, otherwise the purified image will be too close to the adversarial image such that even the adversarial perturbation is preserved in the purified image.

Specifically, we condition the reverse denoising process of the DDPM on the adversarial image $\vx_{\text{adv}}$.
We adapt the reverse denoising distribution $p_{\bm{\theta}}(\vx^{t-1}|\vx^t )$ in Equation~\ref{eqn:reverse_process} to a conditional distribution $p_{\bm{\theta}}(\vx^{t-1}|\vx^t , \vx_{\text{adv}})$.
It is proven by~\citet{sohl2015deep, DBLP:conf/nips/DhariwalN21} that
\begin{align}
    \log p_{\bm{\theta}}(\vx^{t-1}|\vx^t , \vx_{\text{adv}}) = \log [p_{\bm{\theta}}(\vx^{t-1}|\vx^t) p(\vx_{\text{adv}}|\vx^t)] + C_1 \approx \log p(\vz) + C_2, \\
    \text{ where } \vz \sim \gN(\vz;\bm{\mu}_{\bm{\theta}}(\vx^t, t) + \sigma_t^2 \vg, \sigma_t^2\mI), \vg = \nabla_{\vx^t} \log p(\vx_{\text{adv}}|\vx^t), \text{$C_1$ and $C_2$ are some constants} 
    \label{eqn:conditional_guided_reverse}.
\end{align}
In the above equation, $p_{\bm{\theta}}(\vx^{t-1}|\vx^t)$ is the unconditional DDPM that we already have, 
and $p(\vx_{\text{adv}}|\vx^t)$ can be interpreted as
the probability that $\vx^t$ will eventually be denoised to a clean image close to $\vx_{\text{adv}}$.
We propose a heuristic formulation to approximate this probability:
\begin{align}
    p(\vx_{\text{adv}}|\vx^t) = \frac{1}{Z} \exp \left( -s\mathcal{D} \left(\vx^t, \vx^t_{\text{adv}}\right)\right), \text{ where $\mathcal{D}$ is some distance metric,}
\end{align}
$Z$ is a normalization factor, $s$ is a scale factor that controls the magnitude of the guidance, and $\vx^t_{\text{adv}}$ is obtained by diffusing $\vx_{\text{adv}}$ $t$ steps according to Equation~\ref{eqn:xt|x0}.
In this formulation, we requires $\mathcal{D} \left(\vx^t, \vx^t_{\text{adv}}\right)$ to be small in order to increase $p(\vx_{\text{adv}}|\vx^t)$, namely, we encourage the purified image to be close to the adversarial image $\vx_{\text{adv}}$ in the reverse process.

Here we take the logarithm of both sides and then calculate the gradient of them:
\begin{equation}
\label{eqn:p_xadv_xt}
    \begin{split}
        \log p(\vx_{\text{adv}}|\vx^t) = -\log Z - s\mathcal{D} \left(\vx^t, \vx^t_{\text{adv}}\right) ,\\
        \nabla_{\vx^{t}} \log p(\vx_{\text{adv}}|\vx^t)=- s \nabla_{\vx^{t}} \mathcal{D} \left(\vx^t, \vx^t_{\text{adv}}\right).
    \end{split}
\end{equation}

In our experiments, we adopt the mean square error (MSE) or negative structure similarity index measure (-SSIM) as our distance metric $\mathcal{D}$.

After defining $p(\vx_{\text{adv}}|\vx^t)$ and computing its gradients, we can approximate the conditional transition $p_{\bm{\theta}}(\vx^{t-1}|\vx^t , \vx_{\text{adv}})$ with a Gaussian similar to the unconditional transition $p_{\bm{\theta}}(\vx^{t-1}|\vx^t)$, but with its mean shifted by $-s \Sigma \nabla_{\vx^{t}} \mathcal{D} \left(\vx^t, \vx^t_{\text{adv}}\right)$, where $\Sigma$ is the variance of $\vx^t$.
The resulting purification algorithm with the adversarial image as guidance is described in algorithm~\ref{alg2:guidance}. 

\begin{algorithm}[htbp]
	\renewcommand{\algorithmicrequire}{\textbf{Input:}}
	\renewcommand{\algorithmicensure}{\textbf{Output:}}
	\caption{Distance Metric guided diffusion sampling, given a DDPM $\left(\mu_{\theta}\left(x_{t}\right), \Sigma_{\theta}\left(x_{t}\right)\right)$, gradient scale s }
	\label{alg2:guidance}
	\begin{algorithmic}[1]
		\REQUIRE Distance Metric gradient, gradient scale $s$

		\FOR {$i$ $\leftarrow 1$ to $M$}
            \STATE The diffusion process: $\vx^t = \sqrt{\bar{\alpha}_t}(x + v_{perb}) + \sqrt{1-\bar{\alpha}_t}\bm{\epsilon},$
            \FOR{$t \leftarrow T_c$ to $1$}

                \STATE $\mu, \Sigma \leftarrow \mu_{\theta}\left(x_{t}\right), \Sigma_{\theta}\left(x_{t}\right)$
            
            \STATE $x_{t-1} \leftarrow$ sample from $\mathcal{N}\left(\mu-s \Sigma \nabla_{\vx^{t}} \mathcal{D} \left(\vx^t, \vx^t_{\text{adv}}\right), \Sigma\right)$
            \ENDFOR
		\ENDFOR
		\RETURN $x_0$
	\end{algorithmic}
\end{algorithm}


\subsection{Guidance Scale}
The guidance scale $s$ is a critical hyper-parameter that influences purification performance. 
We need it to be large enough to guide the purified image to be close to the adversarial image, and thus the original clean image, 
but at the same time, it should not be too large, otherwise the purified image will be too close to the adversarial image such that even the adversarial perturbation is preserved in the purified image.
We propose to set $s$ as time step $t$-dependent factors $s_t$.
Observing the Equation~\ref{eqn:p_xadv_xt}, we are using $x_{\text{adv}}^t$ as guidance at step $t$, where $\vx^t_{\text{adv}}$ is obtained by diffusing $\vx_{\text{adv}}$ $t$ steps according to Equation~\ref{eqn:xt|x0}.
Adversarial perturbation in $x_{\text{adv}}^t$ is largely submerged and destroyed for large $t$'s, therefore we can bear larger guidance scale $s_t$ without worrying that the purified image will preserve adversarial perturbation from $x_{\text{adv}}^t$.
On the other hand, adversarial perturbation in $x_{\text{adv}}^t$ is largely retained for small $t$'s, therefore we should use small guidance scale $s_t$ to prevent preserving adversarial perturbation in $x_{\text{adv}}^t$.
Therefore, we choose to set $s_t$ proportional to the Gaussian noise magnitude in $x_{\text{adv}}^t$, while disproportional to the adversarial perturbation magnitude in $x_{\text{adv}}^t$.

We know that
\begin{equation}
    \vx_{\text{adv}}^t = \sqrt{\bar{\alpha}_t}(\vx + \bm{\delta}) + \sqrt{1-\bar{\alpha}_t}\bm{\epsilon} = \sqrt{\bar{\alpha}_t}\vx + \sqrt{\bar{\alpha}_t}\bm{\delta} + \sqrt{1-\bar{\alpha}_t}\bm{\epsilon}.
\end{equation}
Each dimension of the adversarial perturbation $\sqrt{\bar{\alpha}_t}\bm{\delta}$ is bounded to $[-\sqrt{\bar{\alpha}_t} \gamma, +\sqrt{\bar{\alpha}_t} \gamma,]$, since we bound the $l_{\infty}$ norm of $\bm{\delta}$ to $\gamma$. 
Each dimension of the Gaussian noise $\sqrt{1-\bar{\alpha}_t}\bm{\epsilon}$ is roughly bounded by $[-3\sqrt{1-\bar{\alpha}_t}, +3\sqrt{1-\bar{\alpha}_t},]$, 
since most samples from the standard Gaussian distribution fall in $[-3,+3]$.
Therefore, we set
\begin{align}
s_t = \frac{3 \sqrt{1-\bar{\alpha}_t}}{\gamma \sqrt{\bar{\alpha}_t}} \cdot a,
\end{align}
where $a$ is an empirically chosen hyper-parameter that depend on the image resolution and the specific distance metric $\mathcal{D}$.

\subsection{DDPM acceleration}
\label{subsec:acceleration}
\citet{nichol2021improved} propose skipping steps in the reverse process to speed up the DDPM generating process. The DDPM does not need to be retrained for this procedure. Their technology is successfully added to our adversarial purification network. We apply this method to the ImageNet dataset purification process, which could accelerate four times faster when we respace T=1000 steps to 250 steps. More experiment data will be attached in supplemental material.

\section{Experiments}

\label{sec:experiment}

The proposed defense method GDMP was evaluated under the strongest existing attacks on bounded threat models and compared with other state-of-the-art defense methods, including adversarial puriﬁcation and adversarial training.
As the purification methods, the Diffusion model and classifier are trained independently on the same dataset. No adversarial training or other training modifications are used for either model.  
For the pre-trained diffusion model, we use models from \citet{DBLP:conf/nips/HoJA20} and \citet{DBLP:conf/nips/DhariwalN21}. We set T=1000 in this paper. For classifiers, three most used architectures is chosen in the following datasets evalutaions: ResNet(\citet{DBLP:conf/cvpr/HeZRS16})50 and ResNet 152, WideResNet(\citet{DBLP:conf/bmvc/ZagoruykoK16})28$\times$10. The datasets we consider in the evaluation are CIFAR10 and ImageNet. 

In this study, adversarial perturbation is considered under $\ell_{\infty}$ norm (i.e., maximum difference for each pixel), with an allowed maximum value of $\gamma$. The value of $\gamma$ is relative to the pixel intensity scale of 255. 

To assess the effectiveness of defense strategies, we use two metrics: \textit{standard accuracy} and \textit{robust accuracy}. The standard accuracy evaluates the defense method's performance on clean data over the whole test set in each dataset. The robust accuracy metric assesses how well the defense method performs on adversarial examples.

As shown in the Figure~\ref{fig:acc_step}, adding guidance can still maintain the classification accuracy when the number of steps increases and has an obvious increase compared to unguided results. Meanwhile, purifying the adversarial images multiple times is shown to be better than just running for one iteration according to Figure~\ref{fig:acc_step}.
\begin{table}[!tbp]
\centering
\caption{CIFAR-10 results for Preprocessor-blind attacks. The PGD attacks to the classiﬁer is performed at $\ell_{\infty} \gamma$-ball with $\gamma$= 8/255. The diffusion length $T_c=36$ and the iteration number $M=4$. The results borrowed from the references are marked with $\ast$.}
\begin{tabular}{llll}
\hline
\multicolumn{1}{c}{\multirow{2}{*}{Models}} & \multicolumn{2}{l}{Accuracy(\%)} & \multirow{2}{*}{Architrcture} \\
\multicolumn{1}{c}{}                        & Standard         & Robust        &                               \\ \hline
\textbf{Raw WideResNet}                     & 95.10             & 0.00          & WRN-28-10                     \\
GDMP(Witout Guidance)                     & 93.5            & 88.74         & WRN-28-10                     \\
GDMP(Guidance metric: SSIM)                     & 93.5            & \textbf{90.10}         & WRN-28-10                     \\
GDMP(Guidance metric: MSE)                     & 93.5            & \textbf{90.06}         & WRN-28-10                     \\\hline
\multicolumn{4}{l}{Adversarial purication models}                                                              \\
\citet{yoon2021adversarial} $^{\ast}$                           &                  &               &                               \\
~~~~ADP($\sigma$ = 0.1)            & 93.09            & 85.45         & WRN-28-10                     \\
~~~~ADP($\sigma$ = 0.25)           & 86.14            & 80.24         & WRN-28-10                     \\
\citet{DBLP:conf/iclr/HillMZ21}  $^{\ast}$                          & 84.12            & 78.91         & WRN-28-10                     \\
\citet{DBLP:conf/iclr/ShiHM21}  $^{\ast}$                               & 96.93            & 63.10         & WRN-28-10                     \\
\citet{DBLP:conf/nips/DuM19} $^{\ast}$                     & 48.7             & 37.5          & WRN-28-10                     \\
\citet{DBLP:conf/iclr/GrathwohlWJD0S20} $^{\ast}$                     & 75.5             & 23.8          & WRN-28-10                     \\
 \citet{DBLP:conf/iclr/SongKNEK18}  $^{\ast}$                         &                  &               &                               \\
~~~~Natural + PixelCNN                          & 82               & 61            & ResNet-62                     \\
~~~~AT + PixelCNN                               & 90               & 70            & ResNet-62                     \\ \hline
\multicolumn{4}{l}{Adversarial training methods, transfer-based}                                               \\
\citet{madry2018towards}  $^{\ast}$                        & 87.3             & 70.2          & ResNet-56                     \\
\citet{zhang2019theoretically}  $^{\ast}$                         & 84.9             & 72.2          & ResNet-56                    \\ \hline
\end{tabular}

\label{table:PGD_result}
\vspace{-5mm}
\end{table}
\paragraph{Precessor-blind attack on CIFAR10 dataset.}

We first evaluate our defense method on adversarial images from PGD attack with standard classifier. The attacker has full access to the classifier but not the purification model. For this experiment, the classifier network is a 28*10 Wide ResNet (\citet{DBLP:conf/bmvc/ZagoruykoK16} ) Classifier, having 36.5M parameters. Moreover, the diffusion model we choose in the CIFAR10 dataset is \citet{DBLP:conf/nips/HoJA20}'s pre-trained model.
In the diffusion process, we define hyperparameters $\beta_t$  according to a linear schedule. We let $\beta_1=1\times10^{-4}$ and $\beta_T=2\times10^{-2}$. Then, we define $\beta_t = \frac{t-1}{T-1}\cdot(\beta_T-\beta_1), t=1,2,\cdots, T$.  
One evaluation of the entire test set took approximately 5 hours using 1$\times$RTX3090 GPU.

\paragraph{Strong adaptive attacks on CIFAR10 dataset.}
Our puriﬁcation algorithm consists of multiple iterations through neural networks, so might cause obfuscated gradient problems. Hence, we also validate our defense method with strong adaptive attacks, including BPDA (\citet{DBLP:conf/icml/AthalyeC018})
We present our assessment results for powerful adaptive assaults on the CIFAR-10 dataset(\citet{krizhevsky2009learning}. Table \ref{table:BPDA_result} shows the assessment findings for a variety of adaptive attacks. We assume that an attacker is aware of the exact purification length utilized in BPDA, and that the attacks are built appropriately. One evaluation of the entire test set took approximately 2.5 days using 8 $\times$A100 GPUs.
\begin{table}[t]
\centering
\caption{CIFAR-10 results for adaptive attacks at $\ell_{\infty} \gamma$-ball with $\gamma$ = 8/255. The diffusion length $T_c=50$ and the iteration number $M=2$. For comparison, we look at various recently proposed preprocessor-based defense strategies as well as adversarial training methods with white-box attacks. The results borrowed from the references are marked with $\ast$.}
\begin{tabular}{llll}
\hline
\multicolumn{1}{c}{\multirow{2}{*}{Models}} & \multicolumn{2}{l}{Accuracy(\%)} & \multirow{2}{*}{Architrcture} \\
\multicolumn{1}{c}{}                        & Natural         & Robust         &                               \\ \hline
GDMP             &  93.50               &                & WRN-28-10                     \\
~~~~BPDA 40+EOT                              &                 & \textbf{79.83}          & WRN-28-10                     \\
~~~~BPDA 100+EOT                             &                 & \textbf{79.57}          &                  WRN-28-10             \\
~~~~SPSA 1280                              &                 & \textbf{87.44}          &        WRN-28-10                       \\\hline
\multicolumn{3}{l}{Adversarial purication models}                              &                               \\
\citet{yoon2021adversarial}$^{\ast}$ADP($\sigma$ = 0.25)        &    86.14              &               &                               \\
~~~~BPDA 40+ EOT            &             & 70.01         & WRN-28-10                     \\
~~~~BPDA 100+ EOT           &             & 69.71         & WRN-28-10                     \\
~~~~SPSA 1280            &             & 80.80         & WRN-28-10                     \\
\citet{DBLP:conf/iclr/HillMZ21}$^{\ast}$                           &                 &                & WRN-28-10                     \\
~~~~BPDA 50 + EOT                               & 84.12           & 54.9           & WRN-28-10                     \\
\citet{DBLP:conf/iclr/SongKNEK18}$^{\ast}$                     &                 &                &                               \\
~~~~BPDA                                        & 95.00           & 9              & ResNet-62                     \\

\citet{DBLP:conf/iclr/ShiHM21} $^{\ast}$                         &                 &                &                               \\
~~~~Classifier PGD 20                           & 91.89           & 53.58          & WRN-28-10                     \\ \hline
\multicolumn{3}{l}{Adversarial training methods, transfer-based}               &                               \\
\citet{madry2018towards} $^{\ast}$                          & 87.3            & 45.8           & ResNet-18                     \\
\citet{zhang2019theoretically} $^{\ast}$                      & 84.9            & 56.43          & ResNet-18                     \\
  \citet{carmon2019unlabeled} $^{\ast}$                    & 89.67           & 63.1           & WRN-28-10                     \\
\citet{gowal2020uncovering} $^{\ast}$                   & 89.48           & 64.08          & WRN-28-10                     \\ \hline
\end{tabular}

\label{table:BPDA_result}
\vspace{-2mm}
\end{table}

\paragraph{Black-box attack on CIFAR10 dataset.} Even if an attacker does not have access to a model or its gradient with respect to a loss function, a large number of samples can be used to estimate the gradient. SPSA ( \citet{DBLP:conf/icml/UesatoOKO18}) is one of these ways, in which random samples near an input are drawn, and the approximate gradient is calculated using the expected value of gradients approximated using the finite-difference method. To make the assault strong enough, we increased the number of inquiries to 1,280. Our result \ref{table:BPDA_result} shows the GDPM could defend most blakc-box attacks.

\begin{table}[t]
\centering
\caption{ImageNet results for preprocessor-blind attacks and strong adaptive attacks.The PGD attacks to the classfier is performed at $\ell_{\infty}~ \gamma$-ball.  The results borrowed from the references are marked with $^{\ast}$. We run these experiments with the acceleration of respace step = 250 and the length of diffusion model = 45.}
\begin{tabular}{llllll}
\hline
\multicolumn{3}{c}{Models}                                                      & \multicolumn{2}{l}{Accuracy(\%)} & \multirow{2}{*}{Architrcture} \\
\multicolumn{1}{c}{Defense Method}     & Attack Method           & $\ell_{\infty}~ \gamma$     & Standard         & Robust        &                               \\ \hline
     GDMP   & Untargeted PGD-100          & 4/255       & 70.17                 & 68.78              & ResNet-50                    \\
        GDMP   & Untargeted PGD-40          & 4/255       &    73.71               & \textbf{70.94}              & ResNet-152                    \\
     GDMP  & Random taregted PGD-40     & 16/255      &   73.71                &      73.45         & ResNet-152                    \\
      GDMP   & BPDA                    & 16/255      &    73.71               &        68.8       & ResNet-152                    \\ \hline
\multicolumn{3}{l}{Adversarial purication models}                &                  &               &                               \\
\citet{DBLP:conf/nips/BaiMYX21}  $^{\ast}$   & Untargeted PGD-100          & 4/255       &  67.38               &     40.27          &               ResNet-50                  \\
\multicolumn{3}{l}{Adversarial training methods, transfer-based}                &                  &               &                               \\
\citet{DBLP:conf/nips/QinMGKDFDSK19} $^{\ast}$      & Untargeted PGD-40          & 4/255       &    72.70              &         47.00      &               ResNet-152                  \\
\citet{DBLP:conf/cvpr/XieWMYH19} $^{\ast}$
                                        & Random targeted PGD-40     & 16/255      &        ----          & 48.6          &       ResNet-152             \\            \hline
\end{tabular}

\label{table:ImageNet}
\vspace{-5mm}
\end{table}
\paragraph{Attacks on ImageNet dataset.}
We also test our approach on the 1.28 million pictures in 1000 classes ImageNet(\citet{imagenet_cvpr09}) classification dataset. On the 50k ImageNet validation pictures that are adversarially disturbed by the attacker, we measure top-1 classification accuracy. The pre-trained diffusion for ImageNet is from \citet{DBLP:conf/nips/DhariwalN21}, which is trained on ImageNet 256$\times$256. However, our classifiers in ImageNet are ResNet-152 and ResNet-50, in which the default image resolution at 224. To solve the image size conflict, we will do an interpolation operation before and after purification progress. 
We define hyperparameters $\beta_t$ same with settings in CIFAR10 dataset. However, the reverse process variances is learned by training rather than calculating. 
While evaluating the ImageNet dataset, we apply the acceleration mentioned in section~\ref{subsec:acceleration} and set the respace timesteps to 250.
Table \ref{table:ImageNet} shows the result of ImageNet compared to previous state-of-art defend methods.
One evaluation of the ImageNet  took approximately 13 hours using 8 $\times$ A100 GPUs under PGD attack.Besides, we also test ImageNet under strong adaptive attacks to verify the obfuscated gradient problems. The result is listed in table \ref{table:ImageNet}. Because of the computation budget limit, we randomly choose 5000 samples from all validation samples to finish adaptive attack. In this case, it also takes 8 $\times$A100 GPUs 14 hours to finish one evaluation.

\section{Conclusion}
\label{sec:conclusion}
In this paper, we propose to use a guided diffusion model for adversarial purification. The proposed method is shown to have a very good purification effect and is an effective tool for noise removal.
We use attack methods based on PGD, BPDA+EOT and black-box attack on the CIFAR-10 and ImageNet dataset to evaluate our defenses, which outperform the predecessors.
However, our method also has the shortage of taking too much time to finish the purification process. It could be further studied to accelerate this method. 
In addition, more studies are needed to prevent this technique from possible misuse such as image watermark removal that causes copyright infringement.

\bibliographystyle{main}
\bibliography{main}

\newpage
\appendix
\section{Appendix}
\subsection{Software and Hardware Conﬁgurations}
We implemented our code on Python version 3.8 and PyTorch version 1.7 with Linux operating system with Slurm Workload Manager. We run each of our experiments on RTX 3090 GPUs or A100 GPUs. Source code for this work is available at https://github.com/JinyiW/GuidedDiffusionPur.git

\subsection{Acceleration Experiment Results}
We tested the acceleration method proposed in \citet{nichol2021improved}, by skipping steps in the reverse process to speed up the DDPM generating process. For this experiment, we test the purification method to defend the PGD attack on ImageNet without guidance. Table~\ref{tab:acceleration} shows the acceleration experiment results in \ref{subsec:acceleration}. 
\begin{table}[h]
\centering
\caption{The acceleration experiment result.}
\begin{tabular}{|l|l|l|l|l|l|l|}
\hline
Time Respace       & 25      & 50      & 100     & 200     & 250     & 500     \\ \hline
Acceleration ratio & 40$\times$     & 20$\times$     & 10$\times$    & 5$\times$     & 4$\times$    & 2$\times$     \\ \hline
Accuracy           & 66.05\% & 67.08\% & 67.47\% & 67.80\% & 67.77\% & 67.68\% \\ \hline
\end{tabular}
\label{tab:acceleration}
\end{table}


\end{document}